\documentclass[letterpaper, 10 pt, journal, twoside]{IEEEtran}

\usepackage{times}
\usepackage{graphics} %
\usepackage{graphicx}
\usepackage{subfigure}
\usepackage{amsmath,amssymb,amsopn,amstext,amsfonts}
\usepackage{cancel}
\usepackage[space]{cite}
\usepackage{pdfsync}
\usepackage{balance}
\usepackage{color}
\usepackage{mathtools}
\usepackage{bm}

\usepackage{diagbox}
\usepackage{float}
\usepackage{epstopdf}
\usepackage{pifont}
\usepackage{fixltx2e}
\usepackage{amsmath}
\usepackage{multirow}
\usepackage{url}
\usepackage{verbatim}
\usepackage{caption}
\usepackage{adjustbox}
\usepackage{booktabs}
\usepackage{threeparttable}
\usepackage{makecell}
\usepackage{tabularx}
\usepackage{arydshln}
\usepackage{siunitx}
\usepackage[normalem]{ulem}
\usepackage{ulem}
\usepackage{fancyhdr}

\usepackage{algorithm}
\usepackage{ifthen}

\usepackage{algorithmic}

\usepackage{array}
\usepackage{textcomp}
\usepackage{stfloats}
\usepackage{amssymb}

\usepackage{siunitx}
\pdfpxdimen=\dimexpr 1 in/72\relax

\newcommand{\etal}{\textit{et al}.}
\newcommand{\ie}{\textit{i}.\textit{e}.}
\newcommand{\eg}{\textit{e}.\textit{g}.}
\newcommand{\zs}[1]{#1}

\def\MYTITLE{Deep Visual Odometry for Stereo Event Cameras}

\makeatletter
\let\NAT@parse\undefined
\makeatother
\usepackage[pagebackref=false,breaklinks=true,colorlinks,bookmarks=true,bookmarksnumbered=true]{hyperref}

\title{\MYTITLE}
\author{Sheng Zhong$^{\ast}$, Junkai Niu$^{\ast}$, Yi Zhou\textsuperscript{\textdagger}   
\thanks{Manuscript received: May 19, 2025; Accepted: August 29, 2025.}
\thanks{This paper was recommended for publication by Editor Javier Civera upon evaluation of the Associate Editor and Reviewers’ comments. 
This work was supported by the National Key Research and Development Project of China under Grant 2023YFB4706600.}
\thanks{All authors are with the Neuromorphic Automation and Intelligence Lab (NAIL) at School of Artificial Intelligence and Robotics, Hunan University, Changsha, China. }
\thanks{$\ast$ equal contribution; \textdagger ~corresponding author (eeyzhou@hnu.edu.cn).}
\thanks{Digital Object Identifier (DOI): see top of this page.}
}

\newcommand{\unum}[1]{\underline{#1}}
\newcommand{\bnum}[1]{\bfseries #1}
\newcommand{\novalue}{{\textendash}}

\definecolor{light-gray}{gray}{0.5}

\begin{document}
\setcounter{figure}{-2} %
\makeatletter
\g@addto@macro\@maketitle{
\vspace{1ex}
  \centering
  \includegraphics[width=0.94\linewidth]{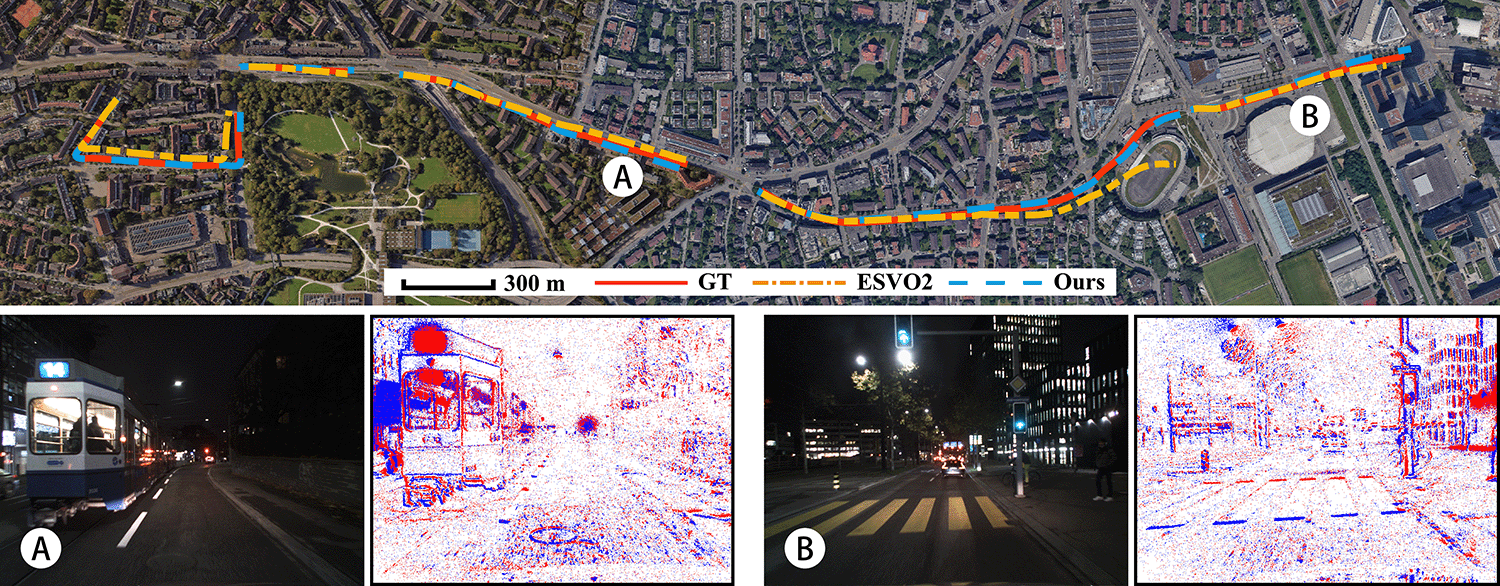}
   \captionof{figure}{
   The proposed system operates in real time on the \emph{dsec\_zurich\_city09}~\cite{Gehrig21ral} sequence, which encompasses large-scale, nighttime HDR scenario at VGA resolution.
   Top: The ground truth trajectory is provided by a LiDAR-IMU-based method~\cite{shan2020lio}  (red) and aligned with Google Maps. 
   Due to the discontinuity of event data, the trajectory is segmented into five parts, with our method (blue) achieving excellent performance on each segment,  outperforming the results of ESVO2~\cite{niu2025esvo2} (yellow).
   Bottom: The scene images and corresponding accumulated event maps (both for visualization purpose only).
   }
   \vspace{-1em}
\label{fig:eye catcher}

\vspace{-3ex}
}
\makeatother
\maketitle

\begin{abstract}
Event-based cameras are bio-inspired sensors with pixels that independently and asynchronously respond to brightness changes at microsecond resolution, offering the potential to handle state estimation tasks involving motion blur and high dynamic range (HDR) illumination conditions.
However, the versatility of event-based visual odometry (VO) relying on handcrafted data association (either direct or indirect methods) is still unreliable, especially in field robot applications under low-light HDR conditions, where the dynamic range can be enormous and the signal-to-noise ratio is spatially-and-temporally varying.
Leveraging deep neural networks offers new possibilities for overcoming these challenges.
In this paper, we propose a learning-based stereo event visual odometry.
Building upon Deep Event Visual Odometry (DEVO), our system (called Stereo-DEVO) introduces a novel and efficient static-stereo association strategy for sparse depth estimation with almost no additional computational burden.
By integrating it into a tightly coupled bundle adjustment (BA) optimization scheme, and benefiting from the recurrent  network's ability to perform accurate optical flow estimation through voxel-based event representations to establish reliable patch associations, our system achieves high-precision pose estimation in metric scale.
In contrast to the offline performance of DEVO, our system can process event data of \zs{Video Graphics Array} (VGA) resolution in real time.
Extensive evaluations on multiple public real-world datasets and self-collected data justify our system's versatility, \zs{demonstrating superior performance compared to state-of-the-art event-based VO methods.}
\zs{More importantly, our system achieves stable pose estimation even in large-scale nighttime HDR scenarios.}
\end{abstract}

\begin{IEEEkeywords}
SLAM, Localization, Deep Learning for Visual Perception.
\end{IEEEkeywords}
\vspace{-0.5em}
\section*{Multimedia Material}
\noindent Video: {\small \url{https://youtu.be/7UykRsmk3Zc}}\\
Code: {\small\url{https://github.com/NAIL-HNU/SDEVO.git}}
\vspace{-0.5em}

\section{Introduction}
\label{sec: introduction}

\IEEEPARstart{N}{euromorphic} event-based cameras are bio-inspired visual sensors that asynchronously detect and transmit pixel-wise intensity changes, known as events.
Consequently, they eliminate grayscale images and fixed capture rates, thereby reducing power consumption and bandwidth requirements. 
Endowed with microsecond temporal resolution and up to 160 dB dynamic range~\cite{Lichtsteiner08ssc}, event cameras are capable of tackling challenging scenarios that are beyond the reach of standard cameras, such as aggressive motion and/or HDR illumination tracking~\cite{Gallego17pami,Mueggler18tro,Wang24tro} and Simultaneous Localization and Mapping (SLAM)~\cite{Rebecq17ijcv,Rosinol18ral,Hidalgo20threedv,Guo24tro}.

Despite the promising potential of event cameras, their unique output renders them incompatible with existing modules in visual VO/SLAM pipelines designed for standard cameras.
How to establish accurate and stable data associations in event data or event representations is a primary challenge for event-based VO/SLAM systems.
A widely adopted approach involves registering reconstructed local 3D maps to event representation images, commonly referred to as the direct method. 
In this context, Niu \etal \cite{niu2025esvo2} recently propose a stereo event-based VIO system called ESVO2 that achieves real-time performance at VGA resolution and demonstrates excellent performance across multiple event datasets, particularly in large-scale scenarios.
However, ESVO2 still faces several limitations. 
Firstly, the inherent limitation of the direct method makes it difficult to handle aggressive motion scenarios. 
Secondly, it fails to operate effectively in complex nighttime HDR scenes, as the flicker of artificial light sources and the significant variation in light intensity with distance can severely disrupt the photometric consistency assumption and the spatio-temporal consistency constraints built upon it.
As shown in Fig.~\ref{fig:eye catcher}, the excessive non-edge events exceed the capacity of ESVO2.

The introduction of deep networks may address these limitations and fully unlock the potential of event cameras in these challenging scenarios. 
Klenk \etal \cite{klenk2023devo} recently propose a learning-based VO approach using a monocular event camera, called DEVO, demonstrating strong generalization and accuracy across extensive datasets from simulation to real environments. 
However, the scale ambiguity of the monocular system and its fully offline operation mode limits its real-world applications.

To overcome the limitations of DEVO~\cite{klenk2023devo}, we extend it to a stereo VO system that enables online and real-time ego motion estimation in metric scale.
Our system demonstrates superior performance in large-scale and HDR night scenes (see Fig.~\ref{fig:eye catcher}), where existing methods \cite{zhou2021event,chen2023esvio, Ghosh24eccvw, niu2025esvo2} fail to operate or perform poorly.
The contribution of the paper is summarized as follows.

\begin{itemize}
    \item \zs{A} learning-based visual odometry system using a stereo event camera, called ``Stereo-DEVO'', which realizes high-precision ego motion estimation in metric scale.
    This is achieved by incorporating a novel and highly efficient static stereo association method, and integrating it into a tightly-coupled BA optimization (Sec.~\ref{sec:method}).

    \item An online system that can operate in real time on event data of VGA resolution.
    This is achieved by refactoring the DEVO framework \cite{klenk2023devo} on ROS, specifically by improving the processing of raw event data and establishing a reliable inter-process communication mechanism. (Sec.~\ref{subsec:implementation details and computational efficiency}).

    \item A comprehensive evaluation on five publicly available datasets and a comparison against six existing methods (including five event-based methods and an image-aided method), demonstrating state-of-the-art accuracy in ego motion estimation and remarkable cross-scene generalization of the proposed method (Sec.~\ref{sec:evaluation}).
    Our code will be released for further research.
\end{itemize}

\emph{Outline}: The rest of the paper is organized as follows.
First, a literature review is provided in Sec.~\ref{sec:related work}.
Second, we provide a discussion of our method, particularly focusing on the entries listed in the contribution (Sec.~\ref{sec:method}).
Finally, experimental results are provided in Sec.~\ref{sec:evaluation}, 
followed by the conclusion (Sec.~\ref{sec:conclusion}).

\section{Related Work}
\label{sec:related work}
Like its standard-vision counterparts, event-based VO/SLAM can be broadly divided into model-based methods and learning-based methods.
The former estimates the camera trajectory by processing events and explicitly constructing photometric and geometric constraints, while the latter utilizes deep networks to learn implicit information from event streams or event representations for trajectory estimation.

\textbf{Model-based Event VO/SLAM:}
From the perspective of event data processing, existing model-based methods can be mainly categorized into feature-based methods and direct methods.
Leveraging the characteristics of event data, researchers modify the original Harris~\cite{Harris88} and FAST~\cite{Rosten06eccv} methods designed for standard cameras to develop new feature detection strategies, such as event corners~\cite{Alzugaray18ral,Li19iros}.
The geometric constraints provided by these feature can be directly utilized by mature geometric tools such as Perspective-\emph{n}-Point (PnP) methods~\cite{li2012robust}.
Despite the success of some feature-based methods (\eg,~\cite{hadviger2021feature}), event features lack robustness because the motion-dependent nature of event data often results in feature tracking failures during sudden changes in the event camera's motion.
Another mainstream strategy for feature detection and tracking is inspired by the motion compensation method \cite{Gallego18cvpr}, a unified pipeline for event-based model fitting.
This strategy is widely employed in event-based visual-inertial odometrt (VIO) pipelines~\cite{Zhu17cvpr,chen2023esvio},  where inertial measurement unit (IMU) measurements are fused to improve system performance through  keyframe-based nonlinear optimization~\cite{leutenegger2013keyframe}.

Unlike feature-based methods, direct methods refer to those that directly process either events or raw pixel information via some event representation.
Based on the constant-brightness assumption in logarithmic scale, Kim \etal \cite{Kim16eccv} propose the first direct method, consisting of three interleaved probabilistic filters to jointly estimate the camera pose, 3D map of the scene, and image intensity.
To justify that intensity information recovery is not needed, Rebecq \etal \cite{Rebecq17ral} propose a geometric approach called EVO, which estimates the camera pose through 3D-2D registration, aligning the local 3D map reconstructed by EMVS~\cite{Rebecq17ijcv} to the 2D event locations on the image plane.
Recently, based on MC-EMVS~\cite{Ghosh22aisy}, the successive work of EMVS, Ghosh \etal \cite{Ghosh24eccvw} employ a similar alignment strategy to develop a more accurate and efficient parallel tracking and mapping system.
Unlike the aforementioned methods, Zhou \etal \cite{zhou2021event} propose the first stereo event-based VO pipeline, which exploits spatio-temporal consistency of the events across the image planes for mapping and localization. 
Building upon this pipeline, \cite{niu2024imu, niu2025esvo2} incorporate IMU measurements and improve event representations to enhance system performance. 
Among these, ESVO2~\cite{niu2025esvo2} achieves state-of-the-art performance in accuracy and efficiency across multiple event datasets by an efficient edge pixel sampling strategy and a tightly-coupled visual-inertial optimization.
However, ESVO2 still suffers from limitations in robustness and generalization, particularly when dealing with aggressive motion and nighttime HDR scenarios.  
These challenges stem from both the inherent limitation of direct methods and the excessive non-edge events in nighttime scenes that compromise spatio-temporal consistency. 
The introduction of deep networks offers new possibilities for overcoming these challenges.

\textbf{Learning-based Event VO/SLAM:}
Zhu \etal \cite{zhu2019unsupervised} and Ye \etal \cite{ye2018unsupervised} employ a convolutional neural network (CNN) to predict optical flow, semi-dense depth, and ego-motion using unsupervised learning on the \emph{MVSEC}~\cite{Zhu18ral} dataset.
Despite achieving favorable results on test data, these methods fail to show generalization beyond the training scenarios.
To address this limitation, Klenk \etal \cite{klenk2023devo} extend DPVO~\cite{teed2024deep} to the event modality by employing a novel patch selection mechanism for sparse event data, called DEVO. 
By training on a very large dataset, it demonstrates generalization from simulation to several real-world benchmarks and achieves competitive performance compared to both classical and deep-learning baselines.
However, DEVO only takes a monocular event camera as input, resulting in trajectory with scale ambiguity.
Building modules upon DEVO, Guan \etal \cite{GWPHKU:DEIO} integrates an event-IMU BA to recover scale information and achieves success on certain small-scale datasets, but its lack of robustness in scale estimation leads to poor performance on larger-scale datasets such as \emph{DSEC}~\cite{Gehrig21ral}.

Similar to \cite{GWPHKU:DEIO}, we also aim to eliminate the scale ambiguity of DEVO. However, our system extends it by introducing stereo event data to construct static stereo associations, achieving stable and accurate depth estimation, and integrating it into a tightly coupled BA optimization module to obtain metric-scale pose estimation.
Our method is a learning-based stereo event VO system, demonstrating  generalization and robustness  across a variety of scenarios and achieving superior performance.

\section{Methodology}
\label{sec:method}

\begin{figure*}[t]
    \centering
    \includegraphics[width=0.9\linewidth]{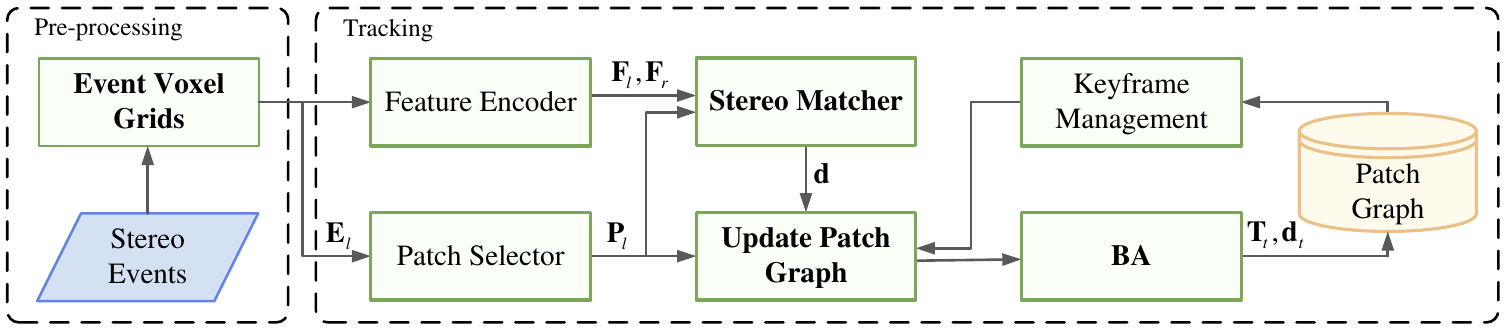}
    \caption{\label{fig:system_overview}
    \emph{Flowchart of the proposed Stereo-DEVO system.} 
    Our system builds on \textit{DEVO}~\cite{klenk2023devo}, with newly added and customized functions highlighted in bold.
    The two main modules (\ie, pre-processing and tracking) run independently online.
    }
    \vspace{-1.5em}
\end{figure*}
\flushbottom

\subsection{System Overview}
\label{subsec:System Overview}
The goal is to deliver an online stereo event-based VO system that can track the 6-DoF motion of the camera in metric scale.
To this end, we build on top of \textit{DEVO}~\cite{klenk2023devo} and extend it to an stereo pipeline with additional non-trivial efforts, as shown in Fig.~\ref{fig:system_overview}.
The pre-processing module receives as input the stereo event stream and performs tri-linear interpolation counting according to the pixel coordinates and timestamps of the events ($e_k=(x_k, y_k, t_k, p_k)$) within a fixed time interval, thereby generating a pair of voxel grids $\mathbf{E}_l,  \mathbf{E}_r\in \mathbb{R}^{H \times W \times 5}$. Here $H$, $W$ denote the sensor's spatial resolution, and $5$ represents that the time interval is divided into five equal segments.

In the tracking module, we reuse the network architecture of \textit{DEVO}, and in particular, the following three functions: \textit{patch selector}, \textit{feature encoder}, and \textit{iterative update operator}, which are incorporated in the \textit{update patch graph}. 
The \textit{patch selector} predicts a score map based on the left voxel grid and uses a pooled multinomial sampling strategy to uniformly extract a fixed number of high-scoring patches $\mathbf{P}_{l}$.
Each patch in $\mathbf{P}_l$ can be denoted as 
\begin{equation}
     \mathbf{P}_i = \left( \mathbf u, \mathbf v, \mathbf 1, \mathbf d \right)^T \ \ \ \  \mathbf{u,v,d} \in \mathbb{R}^{1\times p^2},
\end{equation}
where $\mathbf{u}$ and $\mathbf{v}$ represent the pixel coordinates, while $\mathbf d$ and $p$ ($p=3$ in our implementation) are the inverse depth and the patch size, respectively. 
The \textit{feature encoder} establishes a two-level matching feature pyramid and a context feature map for the left voxel, and extracts only matching features from the right voxel. 
To simplify the notation, we refer to the feature map and the voxel collectively as a frame, denoted by $\mathbf{F}_l$ and $\mathbf{F}_{r}$, respectively.
Based on $\mathbf{P}_l$ and $\mathbf{F}_{l}$, the \textit{iterative update operator} establishes and updates the temporal associations of patches on the left frame within the patch graph by iteratively estimating the optical flow.
Meanwhile, our efficient static stereo association method accurately recovers the depth of the patches and integrate them into the patch graph (Sec.~\ref{subsec:Depth estimation}).
Finally, a tightly-coupled BA module refines the frame poses and patch depths within the sliding window, which is maintained at a fixed size by keyframe management (Sec.~\ref{subsec:Differentiable Bundle Adjustment}).

\subsection{Depth Estimation and Update Patch Graph}
\label{subsec:Depth estimation}
Accurate static stereo associations are fundamental for resolving scale ambiguity through metric depth estimation in a stereo stereo VO system.
Our system follows DEVO's utilization of event sparsity, thereby eliminating the need for dense or semi-dense depth maps as performed in \cite{zhu2019unsupervised, ye2018unsupervised}, which not only require the introduction of additional networks but also significantly increase the computational burden. 
To achieve accurate depth estimation while maintaining system efficiency, we propose a novel and efficient static stereo association method that exploits geometric constraints and the system's existing matching feature pyramids.

Given a patch $\mathbf{P}_i$, let its center coordinates on the first-level matching feature be $(x, y)$.
Since the voxel grids have been stereo rectified, the corresponding matching point on the right feature map must lie between  $(x -D, y)$ and  $(x, y)$,  where $D$ is the maximum disparity.
Let $\mathbf{g}_{l}(x, y)$ and $\mathbf{g}_{{r}}(x, y)$ represent the $p \times p$ patch centered at $(x,y)$ on the first-level matching feature of $\mathbf{F}_l$ and $\mathbf{F}_r$, respectively, then the depth estimation problem can be expressed as:
\begin{equation}
    \delta_{{best}} = \arg \max_{1 \leq \delta \leq D} {\langle \mathbf{g}_{l}(x, y),\ \mathbf{g}_{{r}}(x-\delta, y) \rangle},
\end{equation}
\zs{where $\delta_{best}$ denotes the optimal disparity, corresponding to the static stereo association; and $\langle \cdot \rangle$ represents the dot product, used to measure the similarity between two patches. 
Then the inverse depth can be simply calculated from $\delta_{best}$ and the camera parameters via the operation of triangulation.}
This static stereo association strategy effectively utilizes the information contained in the feature pyramid as a similarity criterion, enabling accurate and robust depth estimation. 
More importantly, it introduces almost no additional computational burden to the system (Sec.~\ref{subsec:implementation details and computational efficiency}), which greatly ensures the system's real-time performance.
Notably, our method performs searches only on first-level feature map, as the GPU's high parallelism allows for faster execution compared to coarse-to-fine searches on two-level feature pyramids without compromising accuracy.

After depth estimation, the newly selected patches with established static stereo associations are incorporated into a fixed-length sliding window along with their corresponding left and right frames, as depicted in Fig.~\ref{fig:backend}. 
The associations between consecutive left frames are constructed and updated by the \textit{iterative update operator}. 
It is noteworthy that the \textit{iterative update operator} is excluded from patch tracking on the right frame due to its reliance on optical flow estimation. 
Unlike consecutive frames, there is no optical flow constraint between the left and right frames at a synchronized timestamp.
Similarly, methods that rely on similarity metrics such as zero-mean normalized cross-correlation (ZNCC) for block matching on voxel grids are not considered, as these methods are prone to mismatches. 
Experiments demonstrate that our static stereo association strategy outperforms these methods in terms of accuracy and robustness (Sec.~\ref{subsec:ablation study}).

\subsection{Bundle Adjustment}
\label{subsec:Differentiable Bundle Adjustment}
\begin{figure}[t]
    \centering
    \includegraphics[width=\linewidth]{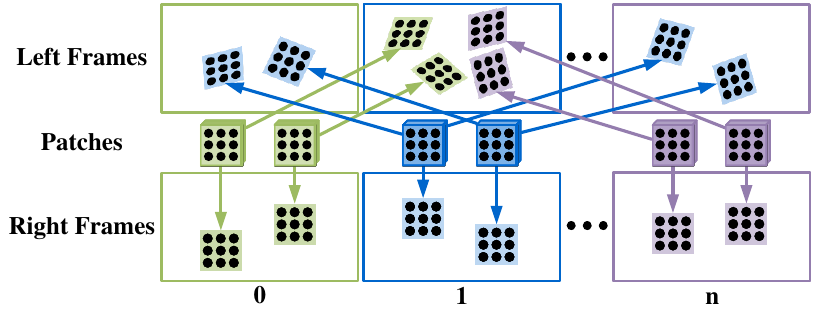}
    \caption{\label{fig:backend}
    \emph{ Patch graph whitin the sliding window.} The patches selected from each frame are connected with the nearby left frames and current right frame.
    }
    \vspace{-2em}
\end{figure}

Building upon the sliding window in Fig.~\ref{fig:backend}, we formulate a tightly-coupled BA optimization problem that accurately and robustly estimates and refines both the camera's metric-scale pose and the patches' depths within the window by simultaneously integrating static and temporal stereo associations.
The state variables in the optimization window are defined as follows:
\begin{equation}
\label{eq:statement}
\begin{split}
  \boldsymbol{\chi} &= [\mathbf{x}_{0},\mathbf{x}_{1} \dots \mathbf{x}_{n }] \\
  \mathbf{x}_{k} &= [\mathbf{R}^{w}_{c_k}, \mathbf{p}^{w}_{c_k}, \lambda^{k}_{0},\lambda^{k}_{1}, \dots \lambda^{k}_{m - 1}], k \in [0, n].
\end{split}
\end{equation}
$\mathbf{x}_{k}$ represents the state of the $k^{th}$ keyframe in the optimization window, including the camera's position and orientation in the world frame, as well as the inverse depth of all patches in that keyframe. $n$ and $m$ denote the number of keyframes and the number of patches per frame, respectively ($n = 10$ and $m = 96$ in our implementation).
We define the optimization problem as minimizing the Mahalanobis norm of the reprojection residuals of patches across all temporal and static stereo associations:
\begin{equation}
    \begin{split}
\label{eq:back-end optimization problem}
   \mathop{\min}\limits_{\boldsymbol{\chi}} 
    \left\{ \sum\limits_{{(i,j)} \in \mathcal{C}} \| r_{\mathcal{C}} 
    (\hat{\boldsymbol z}_{j}^{i}, \boldsymbol{\chi}) \|_{\scalebox{0.5}{$\sum_{ij}$}}^{2}
    + { \sum\limits_{{(l,r)} \in \mathcal{S}}} \| r_{\mathcal{S}} 
    (\hat{\boldsymbol z}_{r}^{l}, \boldsymbol{\chi}) \|_{\scalebox{0.5}{$\sum_{lr}$}}^{2}
    \right\}.
\end{split}
\end{equation}
$r_{\mathcal{C}} (\hat{\boldsymbol z}_{j}^{i}, \boldsymbol{\chi})$ and $r_{\mathcal{S}} (\hat{\boldsymbol z}_{r}^{l}, \boldsymbol{\chi})$ are reprojection residuals for temporal stereo and static stereo, respectively.  
The corresponding sets of associations within the window are denoted by $\mathcal{C}$ and $\mathcal{S}$. 
We solve the optimization problem using two Gauss-Newton iterations to find the optimal camera poses and the inverse depth of corresponding patches.
To improve computational efficiency, the Schur complement is employed to decompose the optimization problem and accelerate the Gauss-Newton iterations.

Each time new frames are added to the window and BA optimization is completed, we compute the optical flow magnitude between the $n-5$ and $n -3$ left frames using the patch graph. 
If this is below than a certain threshold, the $n-4$ frame is removed; otherwise, the oldest frame is discarded. 
These two frames are selected because the patch associations remain relatively stable after multiple updates, reducing the likelihood of misjudgments.

\section{Experiments}
\label{sec:evaluation}

In this section, we provide a comprehensive evaluation of our system.
\zs{It is worth noting that all experiments are conducted without modifying or fine-tuning the network models.}
First, we introduce the publicly available stereo event-based dataset used in the evaluation (Sec.~\ref{subsec:Public Datasets Used}).
Next, we compare the trajectory results of our system with five state-of-the-art event-based VO/VIO algorithms and one image-aided event-based VIO method, demonstrating superior trajectory accuracy compared to existing event-based methods (Sec.~\ref{subsec:Full system evaluation}).
Third, the robustness of our system in both large-scale scenarios (Sec.~\ref{subsec:Evaluation in Large-scale Scenarios}) and aggressive motion scenarios (Sec.~\ref{subsec:Evaluation in aggressive motion Scenarios}) is evaluated using both self-collected data and open-source sequences.
Finally, we investigate the accuracy of the novel static stereo associations (Sec.~\ref{subsec:ablation study}) and provide a computational-complexity analysis of each function in our system (Sec.~\ref{subsec:implementation details and computational efficiency}).

\begin{table}[t]
\centering
\caption{\label{tab:camera parameters}Event camera parameters of the datasets used in the experiments.}
\begin{adjustbox}{max width=\linewidth}
\renewcommand{\arraystretch}{1.2}
\setlength{\tabcolsep}{1.2mm}{
\begin{tabular}{llccc}
\toprule
\textbf{Dataset} & \textbf{Cameras} & \textbf{Resolution} [px] & \textbf{BL} [cm] & \textbf{FOV} [$^\circ$] \\
\midrule

\emph{rpg}~\cite{Zhou18eccv} & DAVIS240C & 240$\times$180 & 14.7 & 62.9\\
\emph{MVSEC}\protect~\cite{Zhu18ral} & DAVIS346 & 346$\times$260 & 10.0 & 74.8\\
\emph{DSEC}\protect~\cite{Gehrig21ral} & Prophesee Gen3.1 & 640$\times$480 & 59.9 & 60.1\\
\emph{VECtor}\protect~\cite{gao2022vector} & Prophesee Gen3 & 640$\times$480 & 17.0 & 67.0\\
\emph{TUM-VIE}\protect~\cite{klenk2021tumvie} & Prophesee Gen4 & 1280$\times$720~ & 11.8 & 65.0\\
\bottomrule
\end{tabular}
}
\end{adjustbox}
\vspace{-3em}
\end{table}

\subsection{Public Datasets Used}
\label{subsec:Public Datasets Used}
To comprehensively evaluate the performance of our system, we use five publicly available stereo event datasets with varying resolutions and scene geometries, including \emph{rpg}~\cite{Zhou18eccv}, \emph{MVSEC}~\cite{Zhu18ral}, \emph{DSEC}~\cite{Gehrig21ral}, \emph{VECtor}~\cite{gao2022vector} and \emph{TUM-VIE}~\cite{klenk2021tumvie}.
The parameters of the event cameras used in each dataset are shown in Tab.~\ref{tab:camera parameters}.

\emph{RPG}, \emph{MVSEC}, \emph{VECtor}, and \emph{TUM-VIE} are all collected in indoor scenes, with four different resolutions ranging from $240 \times 180$ to $1280 \times 720$.
Except for \emph{MVSEC}, which is captured using a stereo event camera mounted on a drone, the other three datasets are collected with a handheld one.
These datasets primarily involve reciprocal motion within the same scene, whereas only \emph{VECtor} provides some sequences featuring long-distance indoor motion. 
\emph{DSEC} consists of a large-scale collection of outdoor driving scenes, featuring a variety of complex scenarios~(e.g., moving object).
Since the \emph{DSEC} dataset does not provide ground truth trajectories, we use the output of the LiDAR algorithm (LIO-SAM~\cite{shan2020lio}) as the ground truth for subsequent evaluation.

\subsection{Full System Evaluation}
\label{subsec:Full system evaluation}

\begin{table*}[!ht]
\begin{minipage}{\textwidth}
\centering
\caption{\label{tab:ate_eval} Relative pose error and Absolute trajectory error (RMS) on
datasets of multiple resolutions [$\mathbf{RPE}$:cm/s, $\mathbf{ATE}$:cm]. Best and second best results are \textbf{highlighted} and \underline{underlined}, respectively.
}
\begin{adjustbox}{max width=\linewidth}
\renewcommand{\arraystretch}{1.11}
\setlength{\tabcolsep}{10pt}
\begin{tabular}{ll*{14}{S[table-format=2.2,table-alignment=right]}}
\toprule
\multirow{2}{*}{\textbf{Dataset}} 
& \multirow{2}{*}{\textbf{Sequence}} 
& \multicolumn{2}{c}{\textbf{ESVO}~\cite{zhou2021event}} 
& \multicolumn{2}{c}{\textbf{ES-PTAM}~\cite{Ghosh24eccvw}} 
& \multicolumn{2}{c}{\textbf{ESIO}~\cite{chen2023esvio}} 
& \multicolumn{2}{c}{\textbf{ESVO2}~\cite{niu2025esvo2}} 
& \multicolumn{2}{c}{\textbf{DEIO}~\cite{GWPHKU:DEIO}} 
& \multicolumn{2}{c}{\textbf{Ours}} \\
\cmidrule(l{1mm}r{1mm}){3-4} 
\cmidrule(l{1mm}r{1mm}){5-6} 
\cmidrule(l{1mm}r{1mm}){7-8} 
\cmidrule(l{1mm}r{1mm}){9-10}
\cmidrule(l{1mm}r{1mm}){11-12}
\cmidrule(l{1mm}r{1mm}){13-14}
{~}&{~}&$\mathbf{RPE}$&$\mathbf{ATE}$&$\mathbf{RPE}$&$\mathbf{ATE}$&$\mathbf{RPE}$&$\mathbf{ATE}$&$\mathbf{RPE}$&$\mathbf{ATE}$&$\mathbf{RPE}$&$\mathbf{ATE}$&$\mathbf{RPE}$&$\mathbf{ATE}$\\
\midrule

\multirow{5}{*}{\emph{rpg}} 
 & \emph{box} & 7.20 & 5.80 &4.222 & \multicolumn{1}{r}{\unum{4.06}} & 20.39 & 11.38 & \multicolumn{1}{r}{\unum{4.18}} & 4.31 & \novalue & \novalue & \bnum{2.8703} & \bnum{1.9212} \\
 & \emph{monitor} & 3.20 & 3.30 & 2.5887 & 2.34 & 13.17 & 7.87 & \multicolumn{1}{r}{\unum{1.69}} & \multicolumn{1}{r}{\unum{2.31}} & \novalue & \novalue & \bnum{1.6547} & \bnum{0.9065} \\
 & \emph{bin}  & 3.10 & 2.80 & \multicolumn{1}{r}{\unum{1.82}} & 2.57 & 12.10 & 7.08 & 2.53 & \multicolumn{1}{r}{\unum{2.27}} & \novalue & \novalue & \bnum{1.5367} & \bnum{1.1604} \\
 & \emph{desk}  & 4.50 & 3.20 & 3.9895 & 2.84 & 6.18 & 3.16 & \multicolumn{1}{r}{\unum{3.53}} & \multicolumn{1}{r}{\unum{1.57}} & \novalue & \novalue & \bnum{3.3301} & \bnum{1.4555} \\
 & \emph{reader}  & 5.60 & 6.60 & \novalue & \novalue & \text{failed} & \text{failed} & \bnum{2.12} & \bnum{2.68} & \novalue & \novalue & \multicolumn{1}{r}{\unum{3.59}} & \multicolumn{1}{r}{\unum{3.78}} \\
\noalign{\vskip 2pt} \hline \noalign{\vskip 2pt}

\multirow{4}{*}{\emph{MVSEC}} 
 & \emph{indoor1\_edited}  & 7.38 & 16.59 & 6.89 & 15.02 & 228.84 & 820.36 & \multicolumn{1}{r}{\unum{5.05}} & 7.63 & 66.6175 & \multicolumn{1}{r}{\unum{7.52}} & \bnum{3.7270} & \bnum{3.7561} \\
 & \emph{indoor2\_edited}  & 7.39 & 14.94 & \novalue & \novalue & 134.48 & 417.85 & \multicolumn{1}{r}{\unum{6.12}} & 10.05 & 53.9232 & \bnum{6.9600} & \bnum{5.7323} & \multicolumn{1}{r}{\unum{8.00}} \\
 & \emph{indoor3\_edited}  & 5.97 & 10.03 & \novalue & \novalue & \text{failed} & \text{failed} & \bnum{4.75} & \multicolumn{1}{r}{\unum{7.35}} & 37.5484 & 29.7524 & \multicolumn{1}{r}{\unum{5.18}} & \bnum{5.5708} \\
 & \emph{indoor4\_edited}  & \text{failed} & \text{failed} & \novalue & \novalue & 181.03 & 173.51 & \bnum{10.36} & \multicolumn{1}{r}{\unum{5.59}} & 122.4763 & 12.7118 & \multicolumn{1}{r}{\unum{14.15}} & \bnum{4.5673} \\ 
\noalign{\vskip 2pt} \hline \noalign{\vskip 2pt}

\multirow{8}{*}{\emph{DSEC}} 
 & \emph{city04\_a}  & 69.14 & 370.32 & 26.9175 & 131.62 & 187.03 & 940.80 & \bnum{16.98} & \bnum{56.17} & 641.2691 & 184.5119 & \multicolumn{1}{r}{\unum{23.70}} & \multicolumn{1}{r}{\unum{96.86}} \\
 & \emph{city04\_b}  & 32.84 & 115.56 & 26.4768 & \multicolumn{1}{r}{\unum{29.02}} & 188.23 & 434.87 & \multicolumn{1}{r}{\unum{21.70}} & 73.83 & 726.4371 & 116.4374 & \bnum{13.5573} & \bnum{17.6544} \\
 & \emph{city04\_c}  & 82.03 & 932.84 & 86.2120 & 1184.37 & 145.64 & 1153.69 & \multicolumn{1}{r}{\unum{39.12}} & \multicolumn{1}{r}{\unum{508.71}} & 1501.4694 & 9128.7281 & \bnum{34.3875} & \bnum{481.6170} \\
 & \emph{city04\_d}  & 190.67 & 2676.11 & 68.0217 & 1053.87 & 540.98 & 6822.53 & \bnum{18.00} & \multicolumn{1}{r}{\unum{546.58}} & 1939.1500 & 10970.0328 & \multicolumn{1}{r}{\unum{31.18}} & \bnum{406.2032} \\
  & \emph{city09\_a}  & 271.62 & 2433.91 & \text{failed} & \text{failed} & \text{failed} & \text{failed} & \multicolumn{1}{r}{\unum{80.86}} & \multicolumn{1}{r}{\unum{1183.04}} & \novalue & \novalue & \bnum{15.73} & \bnum{196.9733} \\
 & \emph{city09\_b}  & 138.331 & 742.19 & 68.9241 & 221.3497 & 561.4102 & 2887.84 & \bnum{32.10} & \bnum{107.4839} & \novalue & \novalue & \multicolumn{1}{r}{\unum{55.00}} & \multicolumn{1}{r}{\unum{267.80}} \\
  & \emph{city09\_c}  & 417.001 & 2962.40 & 1998.860 & 3673.7955 & \text{failed} & \text{failed} & \unum{361.56} & \multicolumn{1}{r}{\unum{1648.30}} & \novalue & \novalue & \bnum{273.98} & \bnum{931.15} \\
 & \emph{city09\_d}  & 280.4301 & 5398.4783 & \text{failed} & \text{failed} & 105.6194 & \multicolumn{1}{r}{\unum{1766.16}} & \multicolumn{1}{r}{\unum{91.51}} & 1920.36 & \novalue & \novalue & \bnum{39.38} & \bnum{625.81} \\
  & \emph{city09\_e}  & 339.240 & 3345.8559 & 744.3031 & 1480.7890 & 801.4313 & 4200.9964 & \unum{125.22} & \multicolumn{1}{r}{\unum{1075.09}} & \novalue & \novalue & \bnum{38.56} & \bnum{246.26} \\
 & \emph{city11\_a}  & 56.82 & 366.22 & 25.14 & 117.86 & 54.53 & 107.36 & \multicolumn{1}{r}{\unum{12.94}} & \multicolumn{1}{r}{\unum{48.77}} & \novalue & \novalue & \bnum{10.3164} & \bnum{47.5576} \\
 & \emph{city11\_b}  & 188.97 & 3241.69 & \text{failed} & \text{failed} & 53.55 & \multicolumn{1}{r}{\unum{300.14}} & \bnum{14.83} & 441.79  & \novalue & \novalue & \multicolumn{1}{r}{\unum{15.58}} & \bnum{282.0640} \\ 
\noalign{\vskip 2pt} \hline \noalign{\vskip 2pt}

\multirow{7}{*}{\emph{VECtor}} 
 & \emph{robot\_normal}  & 16.79 & 7.32 & \novalue & \novalue & 3.38 & 5.17 & \multicolumn{1}{r}{\unum{3.16}} & \multicolumn{1}{r}{\unum{4.81}} & \novalue & \novalue & \bnum{1.8433} & \bnum{3.1643} \\
 & \emph{corner\_slow}  & 6.45 & 13.70 & \novalue & \novalue & 7.01 & 2.67 & \multicolumn{1}{r}{\unum{1.91}} & \multicolumn{1}{r}{\unum{2.15}} & 13.2481 & 10.7005 & \bnum{1.4850} & \bnum{1.0596} \\
 & \emph{hdr\_normal}  & 5.55 & 18.40 & \novalue & \novalue & 15.65 & 27.85 & \multicolumn{1}{r}{\unum{3.79}} & \multicolumn{1}{r}{\unum{13.53}} & \novalue & \novalue & \bnum{1.5487} & \bnum{6.7125} \\
 & \emph{sofa\_normal}  & \text{failed} & \text{failed} & \novalue & \novalue & 54.32 & 43.94 & \multicolumn{1}{r}{\unum{13.59}} & \multicolumn{1}{r}{\unum{40.28}} & \novalue & \novalue & \bnum{4.4386} & \bnum{7.1826} \\
 & \emph{desk\_normal}  & 5.12 & 20.81 & \novalue & \novalue & \text{failed} & \text{failed} & \multicolumn{1}{r}{\unum{3.50}} & \multicolumn{1}{r}{\unum{16.47}} & 86.5500 & 254.7586 & \bnum{2.0016} &  \bnum{6.4825} \\ 
 & \emph{corridors\_dolly} & \text{failed} & \text{failed} & \novalue & \novalue & \text{failed} & \text{failed} & \text{failed} & \text{failed} & \multicolumn{1}{r}{\unum{164.53}} & \multicolumn{1}{r}{\unum{516.08}} & \bnum{16.8675} &  \bnum{196.5306} \\ 
 & \emph{units\_dolly} & \text{failed} & \text{failed} & \novalue & \novalue & \text{failed} & \text{failed} & \text{failed} & \text{failed} & \multicolumn{1}{r}{\unum{138.73}} & \multicolumn{1}{r}{\unum{826.73}} & \bnum{32.43} & \bnum{444.5132} \\ 
\noalign{\vskip 2pt} \hline \noalign{\vskip 2pt}

\multirow{5}{*}{\emph{TUM-VIE}} 
 & \emph{1d\_trans}  & 8.4130 & 12.54 & \bnum{0.7062} & \multicolumn{1}{r}{\unum{1.05}} & \text{failed} & \text{failed} & 1.06 & 3.3344 & 24.4931 & 1.8481 & \multicolumn{1}{r}{\unum{0.89}} & \bnum{1.0466} \\
 & \emph{3d\_trans}  & 7.4727 & 17.19 & 4.9384 & 8.53 & \text{failed} & \text{failed} & \multicolumn{1}{r}{\unum{2.32}} & 7.2635 & 26.4340 & \multicolumn{1}{r}{\unum{1.28}} & \bnum{1.7334} & \bnum{1.2603} \\
 & \emph{6d\_trans}  & 14.2761 & 13.46 & 12.0235 & 10.25 & \text{failed} & \text{failed} & 2.49 & \multicolumn{1}{r}{\unum{3.21}} & 26.3078 & \bnum{1.4471} & \bnum{2.5526} & \multicolumn{1}{r}{\unum{1.69}}\\
 & \emph{desk}  & 5.8707 & 12.92 & 2.5250 & 2.5 & \text{failed} & \text{failed} & \multicolumn{1}{r}{\unum{3.41}} & 6.1611 & 30.751 & \bnum{1.4939} & \bnum{2.2351} & \multicolumn{1}{r}{\unum{2.44}} \\
 & \emph{desk2}  & 6.2277 & 4.42 & 4.0370 & 7.20 & \text{failed}  & \text{failed} & \multicolumn{1}{r}{\unum{2.41}} & 4.02 & 33.6503 & \bnum{1.4497} & \bnum{2.3534} & \multicolumn{1}{r}{\unum{1.79}}\\
 \noalign{\vskip 2pt} \hline \noalign{\vskip 2pt}
 \multirow{2}{*}{\emph{Others}} 
 & \emph{hnu\_campus}  & \novalue & 1321.8024 & \novalue & \novalue & \novalue & 1420.4796 & \novalue & \multicolumn{1}{r}{\unum{181.77}} & \novalue & \novalue & \novalue & \bnum{151.8087} \\
 & \emph{drone\_fast}  & \text{failed} & \text{failed} & \novalue & \novalue & 192.1349  & 154.0421 & \text{failed} & \text{failed} & \novalue & \novalue & \bnum{19.3874} & \bnum{13.2887}\\
\bottomrule
\end{tabular}
\end{adjustbox}
\begin{flushleft}
\vspace{-1ex}
{\emph{* ``-'' represents the lack of experimental results of the algorithm (according to the respective publication). RPE evaluation on $hnu\_campus$ is unavailable due to missing pose ground truth.}}
\end{flushleft}
\end{minipage}
\vspace{-2em}
\end{table*}

\begin{table}[!t]
\begin{minipage}{\linewidth}
\centering
\vspace{0.5em}
\caption{\label{tab:eval_ESVIO} Comparison of trajectory results on the \emph{DSEC} and \emph{VECtor} datasets [$\mathbf{RPE}$:cm/s, $\mathbf{ATE}$:cm].
}
\begin{adjustbox}{max width=\linewidth}
\renewcommand{\arraystretch}{1.11}
\setlength{\tabcolsep}{10pt}
\begin{tabular}{ll*{4}{S[table-format=2.2,table-number-alignment=right]}}
\toprule
\multirow{2}{*}{\textbf{Dataset}} 
& \multirow{2}{*}{\textbf{Sequence}} 
& \multicolumn{2}{c}{\textbf{ESVIO}~\cite{chen2023esvio}} 
& \multicolumn{2}{c}{\textbf{Ours}} \\
\cmidrule(l{1mm}r{1mm}){3-4} 
\cmidrule(l{1mm}r{1mm}){5-6} 
{~}&{~}&$\mathbf{RPE}$&$\mathbf{ATE}$&$\mathbf{RPE}$&$\mathbf{ATE}$\\
\midrule

\multirow{8}{*}{\emph{DSEC}} 
 & \emph{city04\_a}  & 130.7342 & 508.168 & \bnum{23.70} & \bnum{96.86} \\
 & \emph{city04\_b}  & 260.0230 & 129.6969 & \bnum{13.5573} & \bnum{17.6544} \\
 & \emph{city04\_c}  & 279.1668 & 2938.0867 & \bnum{34.3875} & \bnum{481.6170} \\
 & \emph{city04\_d}  & 196.2179 & 2217.3928 & \bnum{31.18} & \bnum{406.20} \\
 & \emph{city09\_a}  & 34.9945 & 328.5136 & \bnum{15.73} & \bnum{196.97} \\
 & \emph{city09\_b}  & 856.5347 & 3481.49 & \bnum{55.00} & \bnum{267.80} \\ 
  & \emph{city09\_c}  & 632.14 & 8043.46 & \bnum{273.98} & \bnum{931.15} \\
 & \emph{city09\_d}  & 128.958 & 1677.1298 & \bnum{39.39} & \bnum{564.33} \\ 
  & \emph{city09\_e}  & 54.1118 & 438.1857 & \bnum{36.49} & \bnum{246.26} \\
\noalign{\vskip 2pt} \hline \noalign{\vskip 2pt}
 \multirow{7}{*}{\emph{VECtor}} 
 & \emph{robot\_normal}  & 2.1976 & 4.9481 & \bnum{1.8433} & \bnum{3.1643} \\
 & \emph{corner\_slow}  & 1.6657 & 1.4272 & \bnum{1.4850} & \bnum{1.0596} \\
 & \emph{hdr\_normal}  & \bnum{0.8998} & \bnum{1.9556} & 1.5487 & 6.7125 \\
 & \emph{sofa\_normal}  & 4.5506 & \bnum{5.0641} & \bnum{4.4386} & 7.1826 \\
 & \emph{desk\_normal}  & 2.2353 & \bnum{6.0042} & \bnum{2.0016} &  6.4825 \\
 & \emph{corridors\_dolly}  & 17.3242 & \bnum{92.0396} & \bnum{16.8675} &  196.5306 \\
 & \emph{units\_dolly}  & 101.9937 & 872.0729 & \bnum{32.4276} & \bnum{444.5132} \\ 
\bottomrule
\end{tabular}
\end{adjustbox}
\end{minipage}
\vspace{-1.5em}
\end{table}

To evaluate the camera pose tracking performance of our system, we conduct extensive comparisons with five event-based VO/VIO systems across five stereo event datasets, as well as with an image-aided stereo event VIO algorithm.
Among them, ESVO~\cite{zhou2021event} is the first event-based stereo VO algorithm using a direct method. 
Building upon it, ESVO2~\cite{niu2025esvo2} also utilizes a direct method and incorporates IMU measurements to enhance the system.
ES-PTAM~\cite{Ghosh24eccvw} is the latest event-only direct stereo VO system, differing from ESVO in that it employs Multi-Camera Event-based Multi-View Stereo (MC-EMVS)~\cite{Ghosh22aisy} for mapping.
DEIO~\cite{GWPHKU:DEIO}, similar to our system, is a successive work of DEVO~\cite{klenk2023devo}, incorporating an IMU to estimate metric-scale poses.
We do not compare with DEVO because its poses suffer from scale ambiguity.
ESVIO~\cite{chen2023esvio} is a feature-based VIO, leverages both standard cameras and event cameras.
They also provide an event-only version (denoted by ESIO), and we compare our system with both of them.
All trajectories used for evaluation are obtained from either the raw data or by running the open-sourced systems. 
We use ``-'' to indicate the absence of raw data, while \text{``failed''} denotes that the open-sourced systems fail to run successfully.

\begin{figure*}[!ht]
	\centering
    \vspace{-0em}
    \begin{minipage}{\textwidth}
    \centering
    \includegraphics[width=\linewidth]{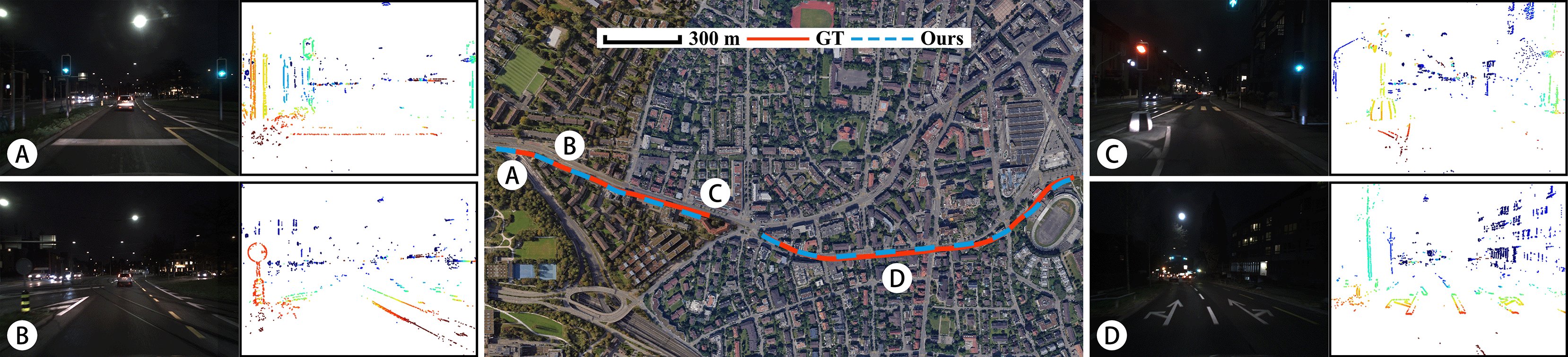}
    \caption{
    Estimated camera trajectories and MC-EMVS reconstructed depth maps on partial sequences of \emph{dsec\_zurich\_city09}.
    Inverse depth maps are color-coded, from red (close) to blue (far) over a white background.
    The scene images are for visualization purposes only.
   }
   \label{fig:mapping_results}
    \vspace{1em}
    \end{minipage}
    
    \begin{minipage}{0.487\textwidth}
    \centering
        \includegraphics[width=\linewidth]{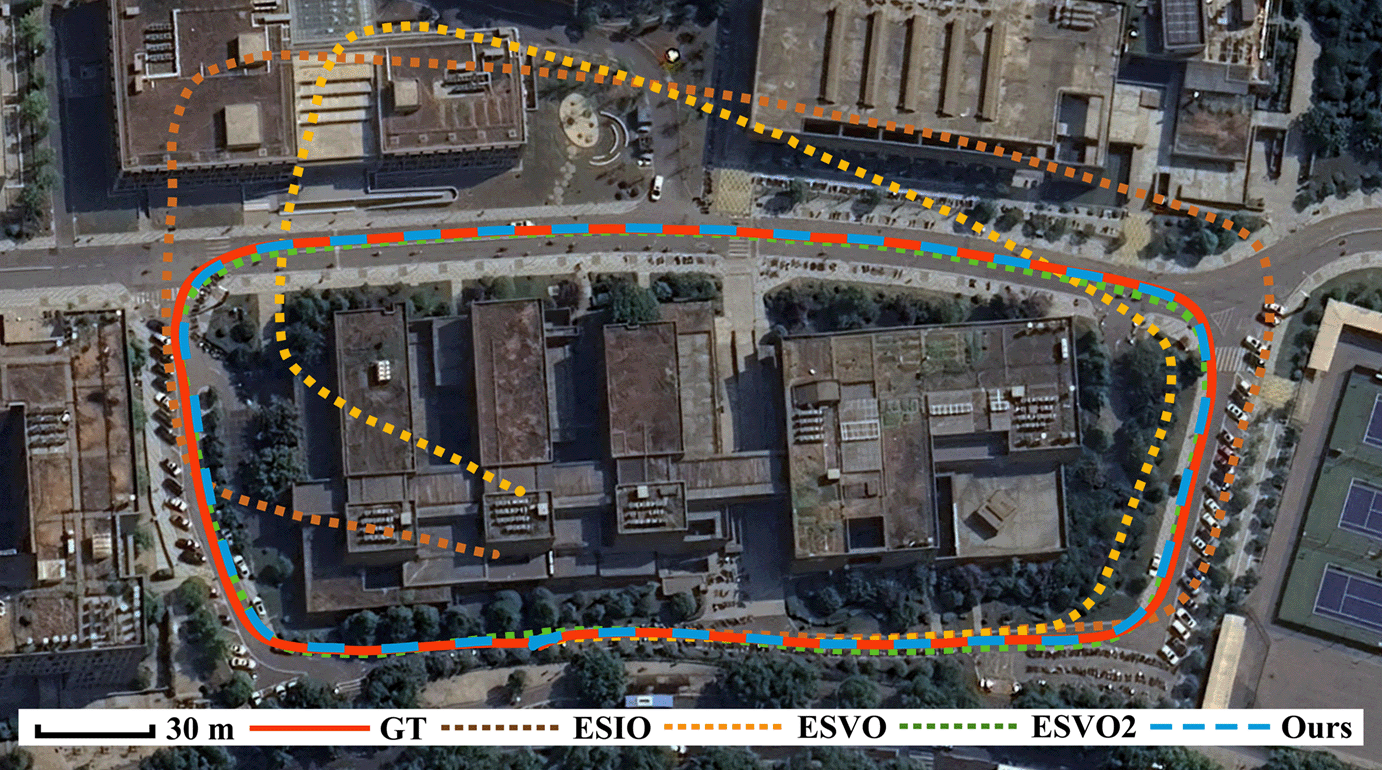}
        \caption{\label{fig:largea_scale}
        \emph{Large-scale loop scenario.}  This scenario is adopted from the \emph{hnu\_campus} sequence of ESVO2~\cite{niu2025esvo2}, where the trajectory ground truth is provided by real-time kinematic (RTK) positioning and aligned with Google Maps.
        }
    \end{minipage}
    \hspace{0.5em}
    \begin{minipage}{0.487\textwidth}
    \centering
        \centering
        \vspace{-1.1em}
        \includegraphics[width=\linewidth]{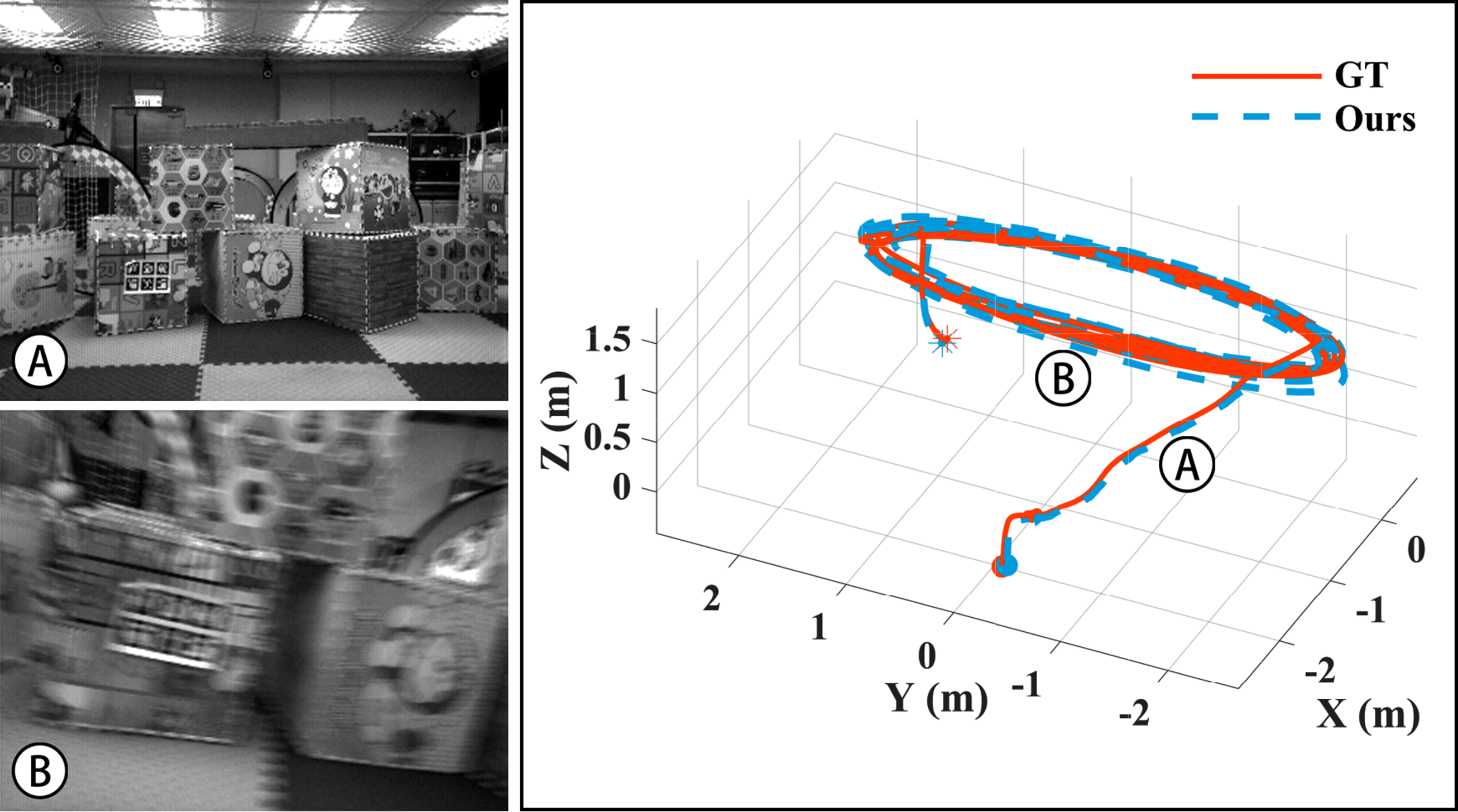}
        \caption{\label{fig:aggressive_motion}
        \zs{\emph{Aggressive motion scenario.} Left column: grayscale images (for visualization only) at two different positions. Right column: trajectory results of our system.}
        }
    \end{minipage}
    \vspace{-1em}
\end{figure*}

To conduct qualitative trajectory evaluation, we employe two standard evaluation metrics: relative pose error (RPE) and absolute trajectory error (ATE) \cite{Sturm12iros}.
The results of all methods are obtained using the EVO tool~\cite{grupp2017evo} and can be found in Tab.~\ref{tab:ate_eval} and Tab.~\ref{tab:eval_ESVIO}.
Our system achieves either the best or the second-best accuracy across all sequences.

Benefiting from the high-precision depth results provided by the multi-view stereo mapping method, ES-PTAM~\cite{Ghosh24eccvw} performs well in small-scale, short-distance motion scenarios. 
In contrast, ESVO2~\cite{niu2025esvo2}, which incorporates a more efficient edge pixel sampling strategy and integrates a tightly-coupled IMU backend, achieves significantly high accuracy in large-scale scenarios like \emph{DSEC}. 
However, due to the inherent limitations of the direct method, ESVO2 struggles to handle scenes with significant optical flow (e.g.~\emph{corridors\_dolly}).
ESIO~\cite{chen2023esvio} is the only feature-based event VIO presented in the table, employing ARC*~\cite{Alzugaray18ral} for feature detection and tracking on motion-compensated event images. 
Notably, ESIO fails to function properly on the \emph{TUM-VIE} dataset, which we attribute to ARC*'s failure to handle excessively blurred edges in high-resolution event images, causing the system to malfunction.
Unlike these methods, our system employs a deep network for patch selection and update on event voxel grids, fully leveraging the optical flow information contained in the events, which enables our system to achieve favorable results on multiple datasets that exhibit diverse geometric characteristics and motion patterns.

Building upon DEVO~\cite{klenk2023devo}, DEIO~\cite{GWPHKU:DEIO} incorporates IMU measurements and integrates it into a tighly-coupled event-IMU BA unit to recover metric-scale poses, achieving outstanding performance on the \emph{TUM-VIE} dataset.
However, it exhibits significant scale errors on other datasets, presumably resulting from the improper weight between IMU and visual constraints.
In comparison, our system eliminates scale ambiguity using static stereo associations, achieving greater robustness. 
Additionally, the initial poses in DEIO's raw data differ from those employed by other algorithms, leading to significant anomalies in its RPE results.

To further demonstrate the performance of our system, we also compare it with ESVIO~\cite{chen2023esvio}, an image-aided event-based stereo VIO system, on the \emph{DSEC} and \emph{VECtor} datasets. 
The results presented in Tab.~\ref{tab:eval_ESVIO} confirm that our system's accuracy is comparable to that of image-based methods.

\subsection{Evaluation in Large-Scale Scenarios }
\label{subsec:Evaluation in Large-scale Scenarios}

We conduct experiments in two large-scale scenarios to evaluate the scalability of our system, as well as its robustness in nighttime HDR environments.
The first scenario uses a sequence called \emph{hnu\_campus} from ESVO2~\cite{niu2025esvo2}, where a stereo-event camera mounted on an electric bicycle captures a closed-loop trajectory returning to the origin in a daytime campus environment, with moving objects introducing dynamic interference.
As demonstrated in Fig.~\ref{fig:largea_scale}, our system's estimated trajectory closely aligns with the ground truth.
The second scenario is the nighttime long-distance driving sequence \emph{zurich\_city09}, provided by \emph{DSEC}~\cite{Gehrig21ral}.
In such a challenging setting, the dim lighting conditions and noise events caused by artificial light sources make it difficult for image-based methods and other event-based approaches to function properly.
However, our system leverages the rich optical flow information from voxel-based event representations and precise patch association mechanisms to achieve state-of-the-art performance. Specifically, we attain a mean absolute trajectory error of 4.52 m over a 2,510 m trajectory (Fig.~\ref{fig:eye catcher}), with detailed quantitative comparisons against baseline methods provided in Tab.~\ref{tab:ate_eval}. 
To further validate trajectory accuracy, we implement 3D scene reconstruction using MC-EMVS~\cite{Ghosh22aisy} on partial sequences of \emph{zurich\_city09}. 
As illustrated in Fig.~\ref{fig:mapping_results}, leveraging the high-precision trajectory results provided by our method, MC-EMVS achieves faithful reconstruction of near-range scenes under challenging nighttime HDR conditions, while far-range scenes remain unrecoverable due to sparse event data.
\begin{table*}[!t]
\centering
\begin{minipage}{0.4\linewidth}
\vspace{-3.2em}
\centering
\caption{\label{tab:ablation_study}Results of ablation study [$\mathbf{ATE}$:cm].
}
\begin{adjustbox}{max width=\linewidth}
\renewcommand{\arraystretch}{1.2}
\setlength{\tabcolsep}{10pt}
\begin{tabular}{l*{3}{S[table-format=2.2,table-number-alignment=right]}}
\toprule
\textbf{Sequence}
& \textbf{Update}
& \textbf{ZNCC}
& \textbf{Ours} \\ 
\midrule
\emph{rpg\_bin}  & 1.6913 & 3.5407 & \bnum{1.16} \\
\emph{mvsec\_indoor1\_edited} & 7.2980 & 6.5715 & \bnum{3.76}\\
\emph{dsec\_city04\_d}  & 528.2243 & 544.1794 & \bnum{406.20} \\
\emph{vector\_robot\_normal} & 48.6729 & 4.0594 & \bnum{3.1643} \\
\emph{vector\_units\_dolly}  & 574.4348 & 491.3981 & \bnum{444.5132} \\ 
\emph{tum\_6d\_trans}  & 385.0752 & 2.4041 & \bnum{1.69} \\
\bottomrule
\end{tabular}
\end{adjustbox}
\end{minipage}
\hspace{2.5em}
\begin{minipage}{0.55\linewidth}
\begin{center}
\renewcommand\arraystretch{1.2}
\caption{Computational performance.[Time: ms]}
\label{computational performance}
\begin{adjustbox}{max width=\linewidth}
\setlength{\tabcolsep}{5pt}{
\begin{tabular}{llrrr}
\toprule
{Node} & {Function} & \textbf{Desktop} & \textbf{Laptop} & \textbf{NVIDIA Jetson Orin NX} \\
\midrule
\multirow{1}{*}{Pre-processing} & Event voxels & 13.52 & 40.41 & 121.40\\
\midrule
\multirow{7}{*}{Tracking} & Feature encoder & 18.39 & 34.13 & 170.51\\
{~} &  Patch selector & 2.92 & 22.28 & 117.60\\
{~} & Depth estimation  & 0.46 & 4.45 & 5.17\\
{~} & Patch graph update  & 12.10 & 92.49 & 433.64\\
{~} & Bundle adjustment  & 1.53 & 3.03 & 17.40\\
{~} & Others  & 27.33 & 42.84 & 201.51\\
\cdashline{2-5}\vspace{-3mm}\\
{~} & Subtotal  & 62.73 & 199.22 & 945.83\\
\bottomrule
\end{tabular}
}
\label{tab:6}
\end{adjustbox}
\end{center}
\end{minipage}
\vspace{-1em}
\end{table*}

\subsection{Evaluation in Aggressive Motion Scenarios}
\label{subsec:Evaluation in aggressive motion Scenarios}

To further assess the performance of our system in aggressive motion scenarios, Fig.~\ref{fig:aggressive_motion} presents the trajectory estimation results in a high-speed indoor sequence called \emph{drone\_fast}, captured by a pair of DAVIS-346 event cameras mounted on a drone.
The drone reaches a maximum linear velocity of 3.9 m/s and performs elliptical motions in the gravity direction. 
Under such aggressive motion, standard cameras exhibit significant motion blur, whereas our system achieves stable pose estimation.
The quantitative results can be found in Tab.~\ref{tab:ate_eval}, where our system yields an absolute trajectory error of only 0.13 m over a 55.4 m trajectory, outperforming other comparison methods.

\subsection{Ablation Study}
\label{subsec:ablation study}

We conduct ablation experiments on sequences from multiple datasets to demonstrate that our static stereo association strategy is more accurate and robust than other methods. 
Specifically, as shown in Tab.~\ref{tab:ablation_study}, ``Ours" refers to using the proposed static stereo association strategy for depth estimation, while ``Update" refers to using the update operator, originally used for temporal stereo association only, to iteratively perform patch tracking twice on the right frame for depth estimation, while ``ZNCC" refers to using ZNCC for block matching along the epipolar lines in the intermediate layer of voxel grids. 
``Update" achieves satisfactory results only in scenarios with small disparities caused by low resolution or large scene depth.
However, when disparities become large, the update operator fails to establish accurate static stereo associations.
Additionally, ``ZNCC" is prone to mismatches, resulting in generally lower trajectory accuracy compared to our method.

\subsection{Computational Efficiency}
\label{subsec:implementation details and computational efficiency}
Unlike DEVO~\cite{klenk2023devo}, which is a fully offline system, Stereo-DEVO is an online system implemented on ROS.
\zs{To assess the computational performance of our system, we evaluate our algorithm on three different platforms: a desktop (Intel Core i7-12700K CPU, RTX 3090 GPU), a laptop (Intel Core i5-12500H CPU, RTX 2050 GPU), and an embedded platform (NVIDIA Jetson Orin NX), with compute precision set to mixed-precision, using the \emph{DSEC} dataset.}
The average run time for each function is presented in Tab.~\ref{computational performance}.

Our system consists of two processes: The pre-processing node, implemented in C++, converts the events into voxel grids, with the entire process executed on the CPU.
The computational bottleneck lies in the tracking node, which is primarily implemented in Python and involves \textit{feature encoder},\textit{ patch selector}, \textit{depth estimation}, \textit{patch graph update}, and \textit{bundle adjustment} on the voxel grids to estimate the trajectory. 
All these operations are accelerated using CUDA on the GPU.
The ``Others'' entry includes operations such as data transfer and keyframe management, with most of the time spent transferring voxel grids from the CPU to the GPU.
\zs{The subtotal runtime of the tracking node confirms that our system can stably operate in real time at 10 Hz on stereo event datasets with VGA resolution on the desktop platform, and at approximately 5 Hz on the laptop platform. 
However, the embedded platform's performance is severely constrained by its 25W power budget, resulting in significantly higher latency.}

\section{Conclusion}
\label{sec:conclusion}

We present a learning-based stereo visual odometry system, which has won 1st place in the M3ED SLAM Challenge at the 5th CVPR Workshop on event-based vision.
The proposed system is built on top of DEVO~\cite{klenk2023devo}, a monocular event-only VO system that leverages a learned patch selector and a pooled multinomial sampling for tracking sparse event patches.
Our system effectively resolves DEVO’s scale ambiguity by integrating stereo event data, establishing robust static stereo correspondences with minimal computational burden. 
These correspondences are then incorporated into a tightly-coupled bundle adjustment optimization framework, thereby enabling precise metric-scale pose estimation.
Furthermore, we enhance DEVO's architecture to transform it from a purely offline system into an online and real-time framework operating on VGA resolution. 
This advancement represents a significant step forward in enabling practical applications of learning-based event visual odometry.
Extensive evaluations on real-world datasets demonstrate our method's superior generalization capability and robustness compared to existing systems. 
Our system particularly excels in large-scale nighttime HDR environments, effectively showcasing the inherent advantages of event cameras in handling such demanding conditions.

\section*{Acknowledgment}
\label{sec:ack}
We thank Dr. Xiuyuan Lu for the help in the experiment.
We also thank Dr. Yi Yu for proofreading.

\bibliographystyle{IEEEtran} %
\bibliography{myBib}

\end{document}